\documentclass[runningheads,a4paper]{llncs}

\usepackage{amssymb}
\usepackage{graphicx}
\usepackage{amsmath}
\usepackage{caption} 
\usepackage{subcaption}
\usepackage{array}
\newcolumntype{M}[1]{>{\raggedright\arraybackslash}m{#1}}
\usepackage[colorinlistoftodos]{todonotes}

\usepackage{algorithm}
\usepackage[noend]{algpseudocode}
\usepackage[title]{appendix}
\usepackage{booktabs}
\usepackage{multirow}

\usepackage{url}
\urldef{\mailsa}\path|{alfred.hofmann, ursula.barth, ingrid.haas, frank.holzwarth,|
\urldef{\mailsb}\path|anna.kramer, leonie.kunz, christine.reiss, nicole.sator,|
\urldef{\mailsc}\path|erika.siebert-cole, peter.strasser, lncs}@springer.com|

\graphicspath{ {./Figures/} }

\begin{document}

\mainmatter  % start of an individual contribution

% first the title is needed
\title{Standard Plane Detection in 3D Fetal Ultrasound Using an Iterative Transformation Network}

% a short form should be given in case it is too long for the running head
\titlerunning{Plane Detection Using Iterative Transformation Network}

% the name(s) of the author(s) follow(s) next
%
% NB: Chinese authors should write their first names(s) in front of
% their surnames. This ensures that the names appear correctly in
% the running heads and the author index.
%
\author{Yuanwei Li\inst{1}\and Bishesh Khanal\inst{2}\and Benjamin Hou\inst{1}\and Amir Alansary\inst{1}\and\\
Juan J. Cerrolaza\inst{1}\and Matthew Sinclair\inst{1}\and Jacqueline Matthew\inst{2}\and\\
Chandni Gupta\inst{2}\and Caroline Knight\inst{2}\and Bernhard Kainz\inst{1}\and Daniel Rueckert\inst{1}}
\authorrunning{Y. Li et al.}
% (feature abused for this document to repeat the title also on left hand pages)

% the affiliations are given next; don't give your e-mail address
% unless you accept that it will be published
\institute{Biomedical Image Analysis Group, Imperial College London, UK
\and
School of Biomedical Engineering \& Imaging Sciences, King's College London, UK
}

%
% NB: a more complex sample for affiliations and the mapping to the
% corresponding authors can be found in the file "llncs.dem"
% (search for the string "\mainmatter" where a contribution starts).
% "llncs.dem" accompanies the document class "llncs.cls".
%

\toctitle{Lecture Notes in Computer Science}
\tocauthor{Authors' Instructions}
\maketitle

\begin{abstract}
Standard scan plane detection in fetal brain ultrasound (US) forms a crucial step in the assessment of fetal development. In clinical settings, this is done by manually manoeuvring a 2D probe to the desired scan plane. With the advent of 3D US, the entire fetal brain volume containing these standard planes can be easily acquired. However, manual standard plane identification in 3D volume is labour-intensive and requires expert knowledge of fetal anatomy. We propose a new Iterative Transformation Network (ITN) for the automatic detection of standard planes in 3D volumes. ITN uses a convolutional neural network to learn the relationship between a 2D plane image and the transformation parameters required to move that plane towards the location/orientation of the standard plane in the 3D volume. During inference, the current plane image is passed iteratively to the network until it converges to the standard plane location. We explore the effect of using different transformation representations as regression outputs of ITN. Under a multi-task learning framework, we introduce additional classification probability outputs to the network to act as confidence measures for the regressed transformation parameters in order to further improve the localisation accuracy. When evaluated on 72 US volumes of fetal brain, our method achieves an error of 3.83mm/12.7$^{\circ}$ and 3.80mm/12.6$^{\circ}$ for the transventricular and transcerebellar planes respectively and takes 0.46s per plane. Source code is publicly available at \url{https://github.com/yuanwei1989/plane-detection}.
%\keywords{Convolutional Neural Network, Classification and Regression, Plane Detection, Fetal Ultrasound}
\end{abstract}

\section{Introduction}
Obstetric ultrasound (US) is conducted as a routine screening examination between 18-24 weeks of gestation. US imaging of the fetal head enables clinicians to assess fetal brain development and detect  growth abnormalities. This requires the careful selection of standard scan planes such as the transventricular (TV) and transcerebellar (TC) plane that contain key anatomical structures \cite{screening2015}. However, it is challenging and time-consuming even for experienced sonographers to manually navigate a 2D US probe to find the correct standard plane. The task is highly operator-dependent and requires a great amount of expertise. With the advent of 3D fetal US, a volume of the entire fetal brain can be acquired quickly with little training. But the problem of locating diagnostically required standard planes for biometric measurements remains. There is a strong need to develop automatic methods for 2D standard plane extraction from 3D volumes to improve clinical workflow efficiency.

\noindent \textbf{\textit{Related work:}}
Recently, deep learning approaches have shown successes in many medical image analysis applications. Several works have applied deep learning techniques to standard plane detection in fetal US~\cite{7974824,chen2015standard,10.1007/978-3-319-24553-9_62,ryou2016automated}. Baumgartner \emph{et al.}~\cite{7974824} use a convolutional neural network (CNN) for categorisation of 13 fetal standard views. Chen \emph{et al.}~\cite{chen2015standard} adopt a CNN-based image classification approach for detecting fetal abdominal standard planes, which they later combined with a recurrent neural network (RNN) that takes into account temporal information \cite{10.1007/978-3-319-24553-9_62}. However, these methods identify standard planes from 2D US videos and not 3D volumes. Ryou \emph{et al.}~\cite{ryou2016automated} attempt to detect fetal head and abdominal planes from 3D fetal US by breaking down the 3D volume into a stack of 2D slices which are then classified as head or abdomen using a CNN.

Plane detection is considered an image classification problem in the above works. In contrast, we approach the plane detection problem by regressing rigid transformation parameters that define the plane position and orientation. There are several works on using CNN to predict transformations. Kendall \emph{et al.}~\cite{kendall2015posenet} introduce PoseNet for regressing 6-DoF camera pose from RGB image with a loss function that uses quaternions to represent rotation. Hou \emph{et al.}~\cite{hou2017predicting} propose  SVRNet for predicting transformation from 2D image to 3D space and use anchor points as a new representation for rigid transformations. These works predict absolute transformation with respect to a known reference coordinate system with one pass of CNN. Our work is different as we use an iterative approach with multiple passes of CNN to predict relative transformation with respect to current plane coordinates, which change at each iteration. Relative transformation is used as our 3D volumes are not aligned to a reference coordinate system.

\noindent \textbf{\textit{Contributions:}}
In this paper, we propose the Iterative Transformation Network (ITN) that uses a CNN to detect standard planes in 3D fetal US. The network learns a mapping between a 2D plane and the transformation required to move that plane towards the standard plane within a 3D volume. Our contributions are threefold: \textbf{(1)} ITN is a general deep learning framework built for 2D plane detection in 3D volumes. The iterative approach regresses transformations that bring the plane closer to the standard plane. This reduces computation cost as ITN selectively samples only a few planes in the 3D volume unlike classification-based methods that require dense sampling~\cite{7974824,chen2015standard,10.1007/978-3-319-24553-9_62,ryou2016automated}. \textbf{(2)} We study the effect on plane detection accuracy using different transformation representations (quaternions, Euler angles, rotation matrix, anchor points) as CNN regression outputs. \textbf{(3)} We improve ITN performance by incorporating additional classification probability outputs as confidence measures of the regressed transformation parameters. \begin{figure}
\centering
\includegraphics[width=\linewidth]{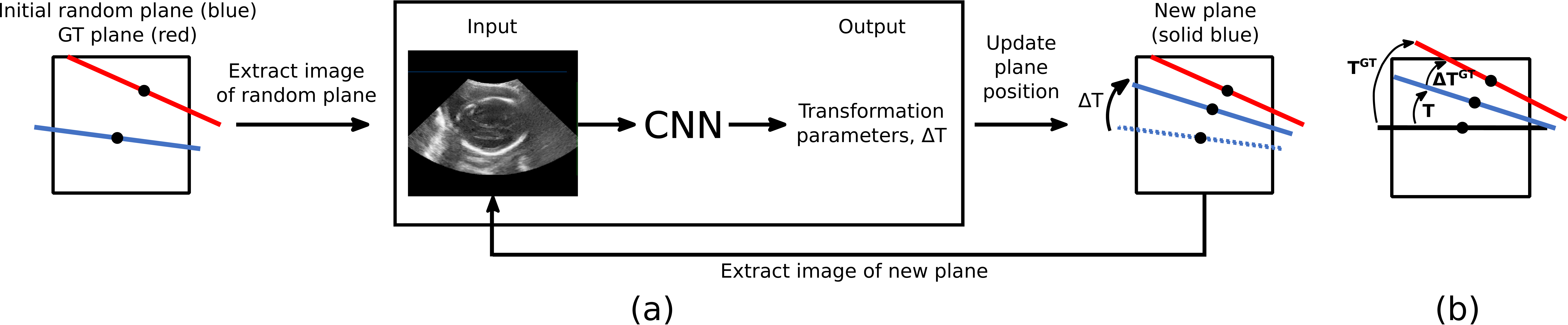}
\caption{(a) Overall plane detection framework using ITN. (b) Composition of transformations. Red: GT plane. Blue: Arbitrary plane. Black: Identity plane.}
\label{fig:pipeline}
\end{figure}
At inference, the classification probabilities are used as confidence scores to yield more accurate localisation. During training, regression and classification outputs are learned in a multi-task learning framework, which improves the generalisation ability of the model and prevents overfitting.

\section{Method}
\noindent\textbf{\textit{Overall Framework:}}
Fig.~\ref{fig:pipeline}a presents the overall ITN framework for plane detection. Given a 3D volume $V$, the goal is to find the ground truth (GT) standard plane (red). Starting with a random plane initialisation (blue), the 2D image of the plane is extracted and input to a CNN which then predicts a 3D transformation $\Delta T$ that will move the plane to a new position closer to the GT plane. The image extracted at the new plane location is then passed to the CNN and the process is repeated until the plane reaches the GT plane.

\noindent\textbf{\textit{Composition of Transformations:}}
Transformation is defined with respect to a reference coordinate system. In Fig.~\ref{fig:pipeline}b, we define an identity plane (black) with origin at the volume centre. $T$ and $T^{GT}$ are defined in the coordinate system of the identity plane and they move the identity plane to the arbitrary plane (blue) and GT plane (red) respectively. $\Delta T^{GT}$ is defined in the coordinate system of the arbitrary plane and $\Delta T^{GT}$ moves the arbitrary plane to the GT plane. Note that our ITN predicts $\Delta T^{GT}$ which is a relative transformation from the point of view of the current plane, and \emph{not} from the identity plane. We compute these transformations from each other using $T^{GT}=T\oplus \Delta T^{GT}$ and $\Delta T^{GT}=T^{GT}\ominus T$ where $\oplus$ and $\ominus$ are the composition and inverse composition operators respectively. The computations defined by the operators are dependent on the choice of the transformation representation.

\noindent\textbf{\textit{Network Training:}}
During training, an arbitrary plane is randomly sampled from a volume $V$ by applying a random transformation $T$ to the identity plane. The corresponding 2D plane image $X$ is then extracted. We define $X=I(V, T, s)$ where $I(\cdot)$ is the plane extraction function and $s$ is the length of the square plane. We sample $T$ such that the plane centre falls in the middle 60\% of $V$ and the rotation of the plane is within an angle of $\pm 45^{\circ}$ about each coordinate axis. This avoids sampling of planes at the edges of the volume where there is no informative image data due to regions falling outside of the US imaging cone. \begin{algorithm}
	\caption{Iterative Inference of Transformation}\label{alg:inference}
	\begin{algorithmic}[1]
		\Procedure{plane transformation}{$V,s,N$}
			\State \text{Initialise random plane with $T_1$}
			\For{$i = 1$ \textbf{to} $N$}
				\State $X_i\gets I(V, T_i, s)$\Comment{Sample  plane image}
				\State $\Delta T\gets \text{CNN}(X_i)$\Comment{CNN predicts relative transformation}
				\State $T_{i+1}\gets T_i \oplus \Delta T$\Comment{Update plane position}
			\EndFor
	      	\State \textbf{return} $T_N$
		\EndProcedure
	\end{algorithmic}
\end{algorithm}
{\renewcommand{\arraystretch}{1.5}%
\begin{table}[]
\centering
\caption{Representations of rigid transformations and their loss functions.}
\label{table:Representations}
\resizebox{0.8\textwidth}{!}{%
\begin{tabular}{|l|l|}
\hline
Representation (Parameter count) & Loss function \\ \hline
Translation $\boldsymbol{t}$ (3) + Quaternion $\boldsymbol{q}$ (4) & $L={ \alpha { \left\| \boldsymbol{t}^{ GT }-\boldsymbol{t} \right\|  }_{ 2 }^{ 2 }+\beta \left\| \boldsymbol{q}^{ GT }-\frac { \boldsymbol{q} }{ \left\| \boldsymbol{q} \right\|  }  \right\|  }_{ 2 }^{ 2 }$ \\ \hline
Translation $\boldsymbol{t}$ (3) + Euler angles $\boldsymbol{\theta}$ (3) & $L={ \alpha { \left\| \boldsymbol{t}^{ GT }-\boldsymbol{t} \right\|  }_{ 2 }^{ 2 }+\beta \left\| \boldsymbol{\theta}^{ GT }-\boldsymbol{\theta} \right\|  }_{ 2 }^{ 2 }$ \\ \hline
Translation $\boldsymbol{t}$ (3) + Rotation matrix $\boldsymbol{R}$ (9) & $L={ \alpha { \left\| \boldsymbol{t}^{ GT }-\boldsymbol{t} \right\|  }_{ 2 }^{ 2 }+\beta \left\| \boldsymbol{R}^{ GT }-\boldsymbol{R} \right\|  }_{ 2 }^{ 2 }$ \\ \hline
Anchor points $(\boldsymbol{A}_1,\boldsymbol{A}_2,\boldsymbol{A}_3)$ (9) & $L=\sum _{ i=1 }^{ 3 }{ { \left\| { \boldsymbol{A} }_{ i }^{ GT }-{ \boldsymbol{A} }_{ i } \right\|  }_{ 2 }^{ 2 } } $ \\ \hline
\end{tabular}}
\end{table}}
A training sample is represented by  $(X, \Delta T^{GT})$ and the training loss function can be formulated as the $L2$ norm of the error between the GT and predicted transformation parameters: $L={ \left\| \Delta { T }^{ GT }-\Delta T \right\|  }_{ 2 }^{ 2 }$

\noindent\textbf{\textit{Network Inference:}}
Algorithm \ref{alg:inference} summarises the steps during network inference to detect a plane. The iterative approach gives rough estimates of the plane in the first few iterations and subsequently makes smaller and more accurate refinements. This coarse-to-fine adjustment improves accuracy and is less susceptible to different initialisations. To improve accuracy and convergence, we repeat Algorithm~\ref{alg:inference} with 5 random plane initialisations per volume and average their final transformations $T_N$ after $N$ iterations.

\noindent\textbf{\textit{Transformation Representations:}}
In ITN, plane transformation $\Delta T$ is rigid, comprising only translation and rotation. We explore the effect of using different transformation representations as the CNN regression outputs (Table~\ref{table:Representations}) since there are few comparative studies that investigate this on deep networks. The first three representations explicitly separate translation and rotation in which rotation is represented by quaternions, Euler angles and rotation matrix respectively. $\alpha$ and $\beta$ are weightings given to the translation and rotation losses. Specifically, anchor points~\cite{hou2017predicting} are defined as the coordinates of three fixed points on the plane (we use: centre, bottom-left and bottom-right corner). The points uniquely and jointly represent any translation and rotation in 3D space. During inference, the predicted values of certain representations need to be constrained to give valid rotation. For instance, quaternions need to be normalised to unit quaternions and rotation matrices need to be orthogonalised. Anchor points need to be converted to valid rotation matrices as described in~\cite{hou2017predicting}.
%{\renewcommand{\arraystretch}{1.5}%
%\begin{table}[]
%\centering
%\caption{Representations of rigid transformations and their loss functions.}
%\label{table:Representations}
%\begin{tabular}{|l|l|l|}
%\hline
%Representation & Number of parameters & Loss function \\ \hline
%Translation $\boldsymbol{t}$ + Quaternion $\boldsymbol{q}$ & 3+4=7 & $L={ \alpha { \left\| \boldsymbol{t}^{ GT }-\boldsymbol{t} \right\|  }_{ 2 }^{ 2 }+\beta \left\| \boldsymbol{q}^{ GT }-\frac { \boldsymbol{q} }{ \left\| \boldsymbol{q} \right\|  }  \right\|  }_{ 2 }^{ 2 }$ \\ \hline
%Translation $\boldsymbol{t}$ + Euler angles $\boldsymbol{\theta}$ & 3+3=6 & $L={ \alpha { \left\| \boldsymbol{t}^{ GT }-\boldsymbol{t} \right\|  }_{ 2 }^{ 2 }+\beta \left\| \boldsymbol{\theta}^{ GT }-\boldsymbol{\theta} \right\|  }_{ 2 }^{ 2 }$ \\ \hline
%Translation $\boldsymbol{t}$ + Rotation matrix $\boldsymbol{R}$ & 3+9=12 & $L={ \alpha { \left\| \boldsymbol{t}^{ GT }-\boldsymbol{t} \right\|  }_{ 2 }^{ 2 }+\beta \left\| \boldsymbol{R}^{ GT }-\boldsymbol{R} \right\|  }_{ 2 }^{ 2 }$ \\ \hline
%Anchor points $(\boldsymbol{A}_1,\boldsymbol{A}_2,\boldsymbol{A}_3)$ & 9 & $L=\sum _{ i=1 }^{ 3 }{ { \left\| { \boldsymbol{A} }_{ i }^{ GT }-{ \boldsymbol{A} }_{ i } \right\|  }_{ 2 }^{ 2 } } $ \\ \hline
%\end{tabular}
%\end{table}}
\begin{algorithm}
	\caption{Compute relative transformation $\Delta T$}\label{alg:delta_T}
	\begin{algorithmic}[1]
		\Procedure{compute Transform}{$\boldsymbol{t},\boldsymbol{q},\boldsymbol{P},\boldsymbol{Q}$}
			\State $\boldsymbol{t}_{new}=\begin{pmatrix} \max { ({ P }_{ { c }_{ 1 }^{ + } } } ,{ P }_{ { c }_{ 1 }^{ - } }){ t }_{ 1 } \\ \max { ({ P }_{ { c }_{ 2 }^{ + } } } ,{ P }_{ { c }_{ 2 }^{ - } }){ t }_{ 2 } \\ \max { ({ P }_{ { c }_{ 3 }^{ + } } } ,{ P }_{ { c }_{ 3 }^{ - } }){ t }_{ 3 } \end{pmatrix}$\Comment{Compute weighted translation}
			\State $Q_{max}=\max{(\boldsymbol{Q})}$\Comment{Compute weighted rotation}
			\If {$Q_{max}=Q_{k^+_x}$ \textbf{OR} $Q_{k^-_x}$}
				\State \text{Convert $\boldsymbol{q}$ to Euler angles $(\theta_x, \theta_y, \theta_z)$ using convention `xyz'}
				\State \text{$\boldsymbol{r}_{new}\gets$Rotation about x-axis with magnitude $Q_{max} \theta_x$}
			\ElsIf {$Q_{max}=Q_{k^+_y}$ \textbf{OR} $Q_{k^-_y}$}
				\State \text{Convert $\boldsymbol{q}$ to Euler angles $(\theta_x, \theta_y, \theta_z)$ using convention `yxz'}
				\State \text{$\boldsymbol{r}_{new}\gets$Rotation about y-axis with magnitude $Q_{max} \theta_y$}
			\ElsIf {$Q_{max}=Q_{k^+_z}$ \textbf{OR} $Q_{k^-_z}$}
				\State \text{Convert $\boldsymbol{q}$ to Euler angles $(\theta_x, \theta_y, \theta_z)$ using convention `zxy'}
				\State \text{$\boldsymbol{r}_{new}\gets$Rotation about z-axis with magnitude $Q_{max} \theta_z$}
			\EndIf
			\State $\Delta T\gets (\boldsymbol{t}_{new}, \boldsymbol{r}_{new})$
	      	\State \textbf{return} $\Delta T$
		\EndProcedure
	\end{algorithmic}
\end{algorithm}

\noindent\textbf{\textit{Classification Probability as Confidence Measure:}}
We further extend our ITN by incorporating classification probability as a confidence measure for the regressed values of translation and rotation. The method can be applied to any transformation representation but we use quaternions since it yields the best results. In addition to the regression outputs $\boldsymbol{t}$ and $\boldsymbol{q}$, the CNN also predicts two classification probability outputs $\boldsymbol{P}$ and $\boldsymbol{Q}$ for translation and rotation respectively. We divide translation into 6 discrete classification categories: positive and negative translation along each coordinate axis. Denoting $c$ as the translation classification label, we have $c\in \{ { c }_{ 1 }^{ + },{ c }_{ 1 }^{ - },{ c }_{ 2 }^{ + },{ c }_{ 2 }^{ - },{ c }_{ 3 }^{ + },{ c }_{ 3 }^{ - }\} $ where ${ c }_{ 1 }^{+}$ is the category representing translation along the positive x-axis. $\boldsymbol{P}$ is then a 6-element vector giving the probability of translation along each axis direction. Similarly, we divide rotation into 6 categories: clockwise and counter-clockwise rotation about each coordinate axis. Denoting $k$ as the rotation classification label, we have $k\in \{ { k }_{ 1 }^{ + },{ k }_{ 1 }^{ - },{ k }_{ 2 }^{ + },{ k }_{ 2 }^{ - },{ k }_{ 3 }^{ + },{ k }_{ 3 }^{ - }\} $ where ${ k }_{ 1 }^{+}$ is the category representing clockwise rotation about the x-axis. $\boldsymbol{Q}$ is then a 6-element vector giving the probability of rotation about each axis.

A training sample is represented by $(X, \boldsymbol{t}^{GT}, \boldsymbol{q}^{GT}, {c^{GT}}, {k^{GT}})$. ${c^{GT}}$ gives the coordinate axis along which the current plane centre has the furthest absolute distance from the GT plane centre. Similarly, ${k^{GT}}$ gives the coordinate axis about which the current plane will rotate the most to reach the GT plane. Appendix A derives the computations of ${c^{GT}}$ and ${k^{GT}}$ during training. 
The overall training loss function can then be written as:
\begin{equation}
L={ \alpha { \left\| \boldsymbol{t}^{ GT }-\boldsymbol{t} \right\|  }_{ 2 }^{ 2 }+\beta \left\| \boldsymbol{q}^{ GT }-\frac { \boldsymbol{q} }{ \left\| \boldsymbol{q} \right\|  }  \right\|  }_{ 2 }^{ 2 } -\gamma \log { { P }_{ c^{ GT } } } -\delta \log { { Q }_{ { k }^{ GT } } } 
\end{equation}
The first and second terms are the $L2$ losses for translation and rotation regression while the third and fourth terms are the cross-entropy losses for translation and rotation classification. $\alpha$, $\beta$, $\gamma$ and $\delta$ are weights given to the losses. 

%During inference, probability ${\boldsymbol{P}}$ and ${\boldsymbol{Q}}$ are used as confidence weighting for the regression outputs $\boldsymbol{t}$ and $\boldsymbol{q}$. This allows the plane to translate and rotate to a greater extent along the more confident axis. Algorithm \ref{alg:delta_T} shows the computation of the final relative transformation $\Delta T$ using the CNN outputs $\boldsymbol{t}$, ${\boldsymbol{P}}$, $\boldsymbol{q}$ and ${\boldsymbol{Q}}$. This is incorporated into Algorithm \ref{alg:inference} which completes the iterative inference. At each iteration, the plane translates a weighted distance specified by ($\boldsymbol{t}$, $\boldsymbol{P}$) and rotates an angle about a most likely axis specified by ($\boldsymbol{q}$, $\boldsymbol{Q}$). 

During inference, the CNN outputs $\boldsymbol{t}$, $\boldsymbol{q}$, ${\boldsymbol{P}}$ and ${\boldsymbol{Q}}$ are combined to compute the relative transformation $\Delta T$ (Algorithm~\ref{alg:delta_T}). For translation, each component of the regressed translation $\boldsymbol{t}$ is weighted by the corresponding probabilities in the vector $\boldsymbol{P}$. For rotation, we only rotate the plane about the most confident rotation axis as predicted by $\boldsymbol{Q}$. In order to determine the magnitude of that rotation, the regressed quaternion $\boldsymbol{q}$ needs to be broken down into Euler angles using the appropriate convention in order to determine the rotation angle about that most confident rotation axis. An Euler angle representation using convention `xyz' means a rotation about x-axis first followed by y-axis and finally z-axis. Hence, ${\boldsymbol{P}}$ and ${\boldsymbol{Q}}$ are used as confidence weighting for $\boldsymbol{t}$ and $\boldsymbol{q}$, allowing the plane to translate and rotate to a greater extent along the more confident axis.

\noindent \textbf{\textit{Network Architecture:}}
ITN utilises a multi-task learning framework for predictions of multiple outputs. The architecture differs according to the number of outputs that the CNN predicts. All our networks comprise 5 convolution layers, each followed by a max-pooling layer. These layers contain shared features for all outputs. After the 5th pooling layer, the network branches into fully-connected layers to learn the specific features for each output. Details of all network architectures are described in Appendix B.

\section{Experiments and Results}
\textbf{\textit{Data and Experiments:}}
ITN is evaluated on 3D US volumes of fetal brain from 72 subjects. For each volume, TV and TC standard planes are manually selected by a clinical expert. 70\% of the dataset is randomly selected for training and the rest 30\% used for testing. All volumes are processed to be isotropic with mean dimensions of 324$\times$207$\times$279 voxels. ITN is implemented using Tensorflow running on a machine with Intel Xeon CPU E5-1630 at 3.70 GHz and one NVIDIA Titan Xp 12GB GPU. We set plane size $s$=225, $N$=10 and $\alpha$=$\beta$=$\gamma$=$\delta$=1. During training, we use a batch size of 64. Weights are initialised randomly from a distribution with zero mean and 0.1 standard deviation. Optimisation is carried out for 100,000 iterations using the Adam algorithm with learning rate=0.001, $\beta_1$=0.9 and $\beta_2$=0.999. The predicted plane is evaluated against the GT using distance between the plane centres ($\delta x$) and rotation angle between the planes ($\delta \theta$). Image similarity of the planes is also measured using peak signal-to-noise ratio (PSNR) and structural similarity (SSIM).

\noindent \textbf{\textit{Results:}}
Table \ref{table:ResultRep} compares the plane detection results when different transformation representations are used by ITN. In general, there is little difference in the translation error. This is because all translation representations are the same, which use the three Cartesian axes except for anchor points which have slightly greater translation error. The rotation errors on TC plane suggest that quaternions are a good representation. Rotation matrices and anchor points over-parameterise rotation and can make network learning more difficult with greater degree of freedom. Since these parameters are not constrained, it is also harder to convert them back into valid rotations during inference. Quaternions have fewer parameters and slightly-off quaternion can still be easily normalised to give valid rotation. Compared to Euler angles, quaternions avoid the problem of gimbal lock. For TV plane, there is little difference in rotation error. \begin{table}[]
\centering
\caption{Evaluation of ITN with different transformation representations for standard plane detection. Results presented as (Mean $\pm$ Standard Deviation).}
\label{table:ResultRep}
\resizebox{\textwidth}{!}{%
\begin{tabular}{|l|c|c|c|c|c|c|c|c|}
\hline
CNN & \multicolumn{4}{c|}{TV plane} & \multicolumn{4}{c|}{TC plane} \\ \cline{2-9}
outputs & $\delta x$ (mm) & $\delta \theta$ ($^{\circ}$) & PSNR & SSIM & $\delta x$ (mm) & $\delta \theta$ ($^{\circ}$) & PSNR & SSIM \\ \hline
$\boldsymbol{t},\boldsymbol{q}$ & 6.29$\pm$5.33 & 17.0$\pm$12.0 & 15.3$\pm$2.0 & 0.375$\pm$0.081 & 6.23$\pm$6.99 & 14.9$\pm$7.5 & 15.5$\pm$2.1 & 0.383$\pm$0.100 \\ \hline
$\boldsymbol{t},\boldsymbol{\theta}$ & 5.69$\pm$5.85 & 17.0$\pm$8.5 & 15.2$\pm$1.7 & 0.372$\pm$0.084 & 7.13$\pm$9.00 & 16.0$\pm$5.9 & 14.6$\pm$2.4 & 0.357$\pm$0.119 \\ \hline
$\boldsymbol{t},\boldsymbol{R}$ & 5.79$\pm$6.10 & 17.7$\pm$11.6 & 15.8$\pm$1.9 & 0.389$\pm$0.091 & 6.39$\pm$7.39 & 17.3$\pm$15.4 & 15.5$\pm$2.4 & 0.385$\pm$0.118 \\ \hline
$\boldsymbol{A}_1,\boldsymbol{A}_2,\boldsymbol{A}_3$ & 6.64$\pm$8.66 & 17.0$\pm$10.4 & 15.9$\pm$2.4 & 0.399$\pm$0.099 & 7.88$\pm$10.0 & 16.3$\pm$12.6 & 15.0$\pm$2.7 & 0.351$\pm$0.124 \\ \hline
\end{tabular}}
\vspace{-0.5cm}
\end{table}
\begin{table}[]
\centering
\caption{Evaluation of ITN with/without confidence probability for standard plane detection. Results presented as (Mean $\pm$ Standard Deviation).}
\label{table:ResultConfidence}
\resizebox{\textwidth}{!}{%
\begin{tabular}{|l|c|c|c|c|c|c|c|c|}
\hline
CNN & \multicolumn{4}{c|}{TV plane} & \multicolumn{4}{c|}{TC plane} \\ \cline{2-9}
outputs & $\delta x$ (mm) & $\delta \theta$ ($^{\circ}$) & PSNR & SSIM & $\delta x$ (mm) & $\delta \theta$ ($^{\circ}$) & PSNR & SSIM \\ \hline
M1: $\boldsymbol{t},\boldsymbol{q}$ & 6.29$\pm$5.33 & 17.0$\pm$12.0 & 15.3$\pm$2.0 & 0.375$\pm$0.081 & 6.23$\pm$6.99 & 14.9$\pm$7.5 & 15.5$\pm$2.1 & 0.383$\pm$0.100 \\ \hline
M2: $\boldsymbol{t},\boldsymbol{q},\boldsymbol{P}$ & 5.14$\pm$5.37 & 16.8$\pm$9.9 & 16.0$\pm$2.1 & 0.408$\pm$0.092 & 5.12$\pm$5.50 & 13.9$\pm$7.1 & 15.8$\pm$2.2 & 0.393$\pm$0.115 \\ \hline
M3: $\boldsymbol{t},\boldsymbol{q},\boldsymbol{Q}$ & 6.07$\pm$6.32 & 14.0$\pm$8.1 & 15.7$\pm$2.5 & 0.399$\pm$0.108 & 7.66$\pm$7.14 & 12.7$\pm$6.0 & 15.5$\pm$3.0 & 0.386$\pm$0.123 \\ \hline
M4: $\boldsymbol{t},\boldsymbol{q},\boldsymbol{P},\boldsymbol{Q}$ & 3.83$\pm$2.10 & 12.7$\pm$7.7 & 16.4$\pm$1.9 & \textbf{0.419$\boldsymbol{\pm}$0.092} & 3.80$\pm$1.85 & 12.6$\pm$6.1 & 16.5$\pm$2.1 & 0.407$\pm$0.110 \\ \hline
M4+: $\boldsymbol{t},\boldsymbol{q},\boldsymbol{P},\boldsymbol{Q}$ & \textbf{3.49$\boldsymbol{\pm}$1.81} & \textbf{10.7$\boldsymbol{\pm}$5.7} & \textbf{16.6$\boldsymbol{\pm}$1.8} & 0.413$\pm$0.082 & \textbf{3.39$\boldsymbol{\pm}$2.13} & \textbf{11.4$\boldsymbol{\pm}$6.3} & \textbf{16.8$\boldsymbol{\pm}$2.1} & \textbf{0.437$\boldsymbol{\pm}$0.110} \\ \hline
\end{tabular}}
\vspace{-0.4cm}
\end{table}
This is because sonographers use the TV plane as a visual reference when acquiring 3D volumes. This causes the TV plane to lie roughly in the central plane of the volume with lower rotation variances, thus making the choice of rotation representation less important. Table \ref{table:ResultConfidence} compares the performance of ITN with/without classification probability outputs. Given a baseline model (M1) that only has regression outputs $\boldsymbol{t}$, $\boldsymbol{q}$, the addition of classification probabilities $\boldsymbol{P}$, $\boldsymbol{Q}$ improves the translation and rotation accuracy respectively (M2-M4). The classification probabilities act as confidence weights for the regression outputs to improve plane detection accuracy. Furthermore, the classification and regression outputs are trained in a multi-task learning fashion, which allows feature sharing and enables more generic features to be learned, thus preventing model overfitting. M1-M4 use one plane image as CNN input. We further improve our results by using three orthogonal plane images instead as this provides more information about the 3D volume (M4+). M4 and M4+ take 0.46s and 1.35s to predict one plane per volume. The supplementary material provides videos showing the update of a randomly initialised plane and its extracted image through 10 inference iterations.

\begin{figure}
\centering
\includegraphics[width=\linewidth]{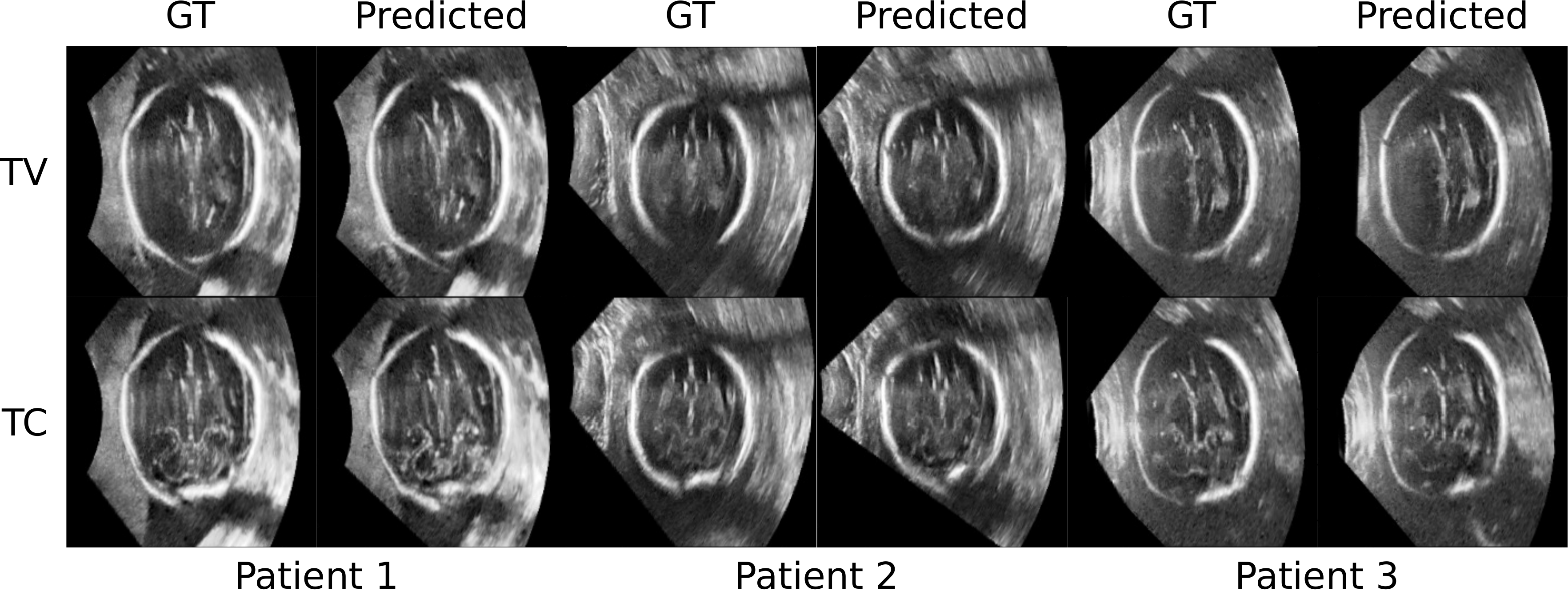}
\caption{Visualisation of GT planes and planes predicted by M4.}
\label{fig:results}
\end{figure}

Fig.~\ref{fig:results} shows a visual comparison between the GT planes and the planes predicted by M4. To evaluate the clinical relevance of the predicted planes, a clinical expert manually measures the head circumference (HC) on both the predicted and GT planes and computes the standard deviation of the measurement error to be 1.05mm (TV) and 1.25mm (TC). This is similar to the intraobserver variability of 2.65mm reported for HC measurements on TC plane \cite{sarris2012intra}. Thus, accurate biometrics can be extracted from our predicted planes.

\section{Conclusion}
We presented ITN, a new approach for standard plane detection in 3D fetal US by using a CNN to regress rigid transformation iteratively. We compare the use of different transformation representations and show quaternions to be a good representation for iterative pose estimation. Additional classification probabilities are learned via multi-task learning which act as confidence weights for the regressed transformation parameters to improve plane detection accuracy. As future work, we are evaluating ITN on other plane detection tasks (\emph{eg.} view plane selection in cardiac MRI). It is also worthwhile to explore new transformation representations and extend ITN to simultaneous detection of multiple planes.

\subsubsection*{Acknowledgments.} Supported by the Wellcome Trust IEH Award [102431]. The authors thank Nvidia Corporation for the donation of a Titan Xp GPU. %integrate after acceptance
\bibliography{References}

\end{document}